\documentclass[conference]{IEEEtran}
\IEEEoverridecommandlockouts
\usepackage{cite}
\usepackage[font=footnotesize]{caption}
\usepackage{amsmath,amssymb,amsfonts}
\usepackage[linesnumbered, ruled, vlined]{algorithm2e}
\usepackage[nolist]{acronym}
\usepackage{multirow}
\usepackage{xcolor}
\usepackage{graphicx}
\usepackage{hyperref}
\hypersetup{colorlinks=true, unicode=true, linkcolor=[rgb]{0.10,0.05,0.67}, citecolor=[rgb]{0.10,0.05,0.67}, filecolor=[rgb]{0.10,0.05,0.67}, urlcolor=[rgb]{0.10,0.05,0.67}}
\usepackage{xcolor}
\usepackage{mathtools}
\usepackage{bbm} 
\newcommand{\greenup}{\textcolor{green}{$\uparrow$}}
\newcommand{\reddown}{\textcolor{red}{$\downarrow$}}
\newtheorem{definition}{Definition}
\newcommand{\argmin}{\operatornamewithlimits{argmin}}
\newcommand{\argmax}{\operatornamewithlimits{argmax}}

\def\BibTeX{{\rm B\kern-.05em{\sc i\kern-.025em b}\kern-.08em
    T\kern-.1667em\lower.7ex\hbox{E}\kern-.125emX}}

\begin{document}

\makeatletter
    \newcommand{\linebreakand}{%
      \end{@IEEEauthorhalign}
      \hfill\mbox{}\par
      \mbox{}\hfill\begin{@IEEEauthorhalign}
    }
    \makeatother
\author{%
\IEEEauthorblockN{Andreas Lutz\textsuperscript{1,2}\thanks{Corresponding author — Email: \texttt{andreas.lutz@charite.de}},\ 
Gabriele Steidl\textsuperscript{3},\ 
Karsten Müller\textsuperscript{2},\ 
Wojciech Samek\textsuperscript{2,4,5}}
\IEEEauthorblockA{\textsuperscript{1}Institut für Künstliche Intelligenz in der Medizin (IKIM), 
Charité – Universitätsmedizin Berlin, Germany}

\IEEEauthorblockA{\textsuperscript{2}Department of Artificial Intelligence, Fraunhofer Heinrich-Hertz-Institute, Berlin, Germany}
\IEEEauthorblockA{\textsuperscript{3}Department of Mathematics, Technische Universität Berlin, Germany}
\IEEEauthorblockA{\textsuperscript{4}Department of Electrical Engineering and Computer Science, Technische Universität Berlin, Germany}
\IEEEauthorblockA{\textsuperscript{5}BIFOLD – Berlin Institute for the Foundations of Learning and Data, Berlin, Germany}
}

\title{Optimizing Federated Learning by Entropy-Based Client Selection}

\maketitle

\begin{abstract}
Although deep learning has revolutionized domains such as natural language processing and computer vision, its dependence on centralized datasets raises serious privacy concerns. Federated learning addresses this issue by enabling multiple clients to collaboratively train a global deep learning model without compromising their data privacy. However, the performance of such a model degrades under label skew, where the label distribution differs between clients.
To overcome this issue, a novel method called FedEntOpt is proposed. In each round, it selects clients to maximize the entropy of the aggregated label distribution, ensuring that the global model is exposed to data from all available classes.
Extensive experiments on multiple benchmark datasets show that the proposed method outperforms several state-of-the-art algorithms by up to 6\% in classification accuracy under standard settings regardless of the model size, while achieving gains of over 30\% in scenarios with low participation rates and client dropout. In addition, FedEntOpt offers the flexibility to be combined with existing algorithms, enhancing their classification accuracy by more than 40\%. Importantly, its performance remains unaffected even when differential privacy is applied.
\end{abstract}
\section{Introduction}
Nowadays, a large amount of data is continuously created from smartphones, desktop computers, and other Internet of
Things (IoT) devices. Approximately 500 million terabytes of data are produced daily, and the volume is expected to double in the next three years \cite{bibliography:data-generation}.
Deep learning models are typically trained on these data to achieve state-of-the-art performance \cite{bibliography:a-survey-on-deep-learning-tools-dealing-with-data-scarcity}.
However, storing all this information in a centralized location for training such models is not possible due to data privacy violations such as the General Data Protection Regulation (GDPR) \cite{bibliography:data-eu-regulation}.
Consequently, data remains fragmented over different nodes, which limits the model performance.
\Ac{fl} addresses this issue by letting clients train a shared deep learning model on their local data, exchanging only model parameters for aggregation, and thus preserving privacy \cite{bibliography:communication-efficient-learning-of-deep-networks-from-decentralized-data}. 
However, simply averaging the updated model parameters can significantly decrease the performance of the global model due to heterogeneous data distributions among clients.
The worst case is label distribution skew, where class proportions vary between clients, resulting in up to a 26\% drop in classification accuracy compared to the case where the label distribution among clients is homogeneous \cite{bibliography:federated-learning-with-label-distribution-skew-via-logits-calibration}. Furthermore, label skew is often encountered in practice, as individual clients acquire their data independently. For instance, hospitals in urban areas may have more images of respiratory diseases because their patients are exposed to higher levels of air pollution compared to patients in rural areas.
To mitigate the impact of label skew, prior work constrains local updates to keep models close to the global one by using a modified loss function \cite{bibliography:fedrs-federated-learning-with-restricted-softmax-for-label-distribution-non-iid-data}, incorporating contrastive learning \cite{bibliography:model-contrastive-federated-learning} or introducing control variates \cite{bibliography:scaffold-stochastic-controlled-averaging-for-federated-learning}. Other works try to match the global label distribution by employing weighting schemes to aggregate locally acquired updates \cite{bibliography:label-aware-aggregation-for-improved-federated-learning}, cluster clients with similar label distribution \cite{bibliography:exploiting-label-skews-in-federated-learning-with-model-concatenation} or choose a subset of clients based on a specific metric \cite{bibliography:fedcor-correlation-based-active-client-selection-strategy-for-heterogeneous-federated-learning,
bibliography:client-selection-for-federated-learning-with-non-iid-data-in-mobile-edge-computing}. 
There also exist entropy-based approaches that aim for a similar goal \cite{bibliography:entropy-to-mitigate-non-iid-data-problem-on-federated-learning-for-the-edge-intelligence-environment,bibliography:fedentropy-efficient-device-grouping-for-federated-learning-using-maximum-entropy-judgment}. Nevertheless, all of the mentioned methods do not address the issue that the chosen client subset may still lack samples for certain classes, biasing the global model toward over-represented classes and pushing it into higher loss regions on its loss surface. \cite{bibliography:understanding-and-improving-model-averaging-in-federated-learning-on-heterogeneous-data}. 
To improve on this, we propose an entropy-based client selection strategy, which optimizes the selection of a subset of clients that is representative of the overall label distribution. Therefore, our research makes the following contributions:
\begin{itemize}
    \item We introduce FederatedEntropyOptimization (FedEntOpt), which iteratively selects clients by maximizing the entropy of the aggregated label distribution, effectively approximating the global label distribution.
    
    \item We evaluate FedEntOpt on several datasets against other state-of-the-art algorithms, showing superior classification accuracy regardless of the model size.
    
    \item We empirically demonstrate that our method is robust under different client participation rates, especially in low-participation scenarios, thereby reducing communication overhead compared to existing algorithms.
    
    \item We illustrate the robustness of FedEntOpt against client dropout and straggling, keeping high accuracy even when many clients are unavailable or send partial updates.
    
    \item We show that combining FedEntOpt with other algorithms improves their classification accuracy.
    
    \item We demonstrate the robustness of FedEntOpt when incorporating differential privacy, with negligible or even positive impact on accuracy.
\end{itemize}

\section{Related Work}

\subsubsection{Federated Learning with Heterogeneous Data}
FedAvg \cite{bibliography:communication-efficient-learning-of-deep-networks-from-decentralized-data} laid the groundwork for \ac{fl}, which averages transmitted model parameters weighted by the dataset size of participating clients. However, it converges slowly and loses accuracy when local data distributions vary \cite{bibliography:on-the-convergence-of-fedavg-on-non-iid-data}.
FedProx \cite{bibliography:federated-optimization-in-heterogeneous-networks} improves convergence by incorporating a proximal term into each client’s local loss  
function, but fails to account for the difference between client models.
SCAFFOLD \cite{bibliography:scaffold-stochastic-controlled-averaging-for-federated-learning} addresses this by introducing control variates to correct for client drift at the cost of additional communication overhead.
FedNova \cite{bibliography:tackling-the-objective-inconsistency-problem-in-heterogeneous-federated-optimization} addresses heterogeneity during aggregation by normalizing client updates to align the optimization of different local objectives.
All of these methods have been shown to degrade under label skew, as aggregation alone cannot control the global label distribution among selected clients \cite{bibliography:aggregation}. FedEntOpt addresses this issue by optimizing client selection directly.

\subsubsection{Federated Learning with Label Skew}
To handle label skew, FedRS \cite{bibliography:fedrs-federated-learning-with-restricted-softmax-for-label-distribution-non-iid-data} relies on a loss function that constrains weight updates for missing labels. FedLC \cite{bibliography:federated-learning-with-label-distribution-skew-via-logits-calibration} improves on this by calibrating the logits according to the local label distribution.
FedConcat \cite{bibliography:exploiting-label-skews-in-federated-learning-with-model-concatenation} clusters clients by label distribution, applies FedAvg within each cluster, and then concatenates the resulting model parameters to form the global model. However, extreme label imbalance requires several clusters, leading to increased computational and storage demands. All of the aforementioned methods lack the consideration of the aggregated label distribution across all selected clients in each communication round, which can lead to suboptimal performance compared to our entropy-based selection method. Another work \cite{bibliography:performance-enhancement-in-federated-learning-by-reducing-class-imbalance-of-non-iid-data} utilizes label information transmitted by each client before actual training. However, their optimization method consistently produces the same subset of clients and excludes large portions of local data, leading to compromised model performance, as we will demonstrate in the experiment section.

\subsubsection{Federated Learning with Entropy}
Several works employ entropy-based approaches to mitigate heterogeneity in \ac{fl}.
Fed\-Ent~\cite{bibliography:entropy-theory} optimizes an entropy term over local models to derive an adaptive learning rate for each client. 
However, it does not improve class coverage during aggregation, i.e., the extent to which the data of the selected clients collectively span the label space, since FedEnt only optimizes how clients update, not which clients are chosen. By design, FedEntOpt optimizes client selection to maximize coverage.
Another approach, EntropicFL \cite{bibliography:entropicfl-efficient-federated-learning-via-data-entropy-and-model-divergence}, leverages both data entropy and model divergence to guide which clients participate and whether their updates should be transmitted, reducing communication overhead at the cost of compromised performance comparable to FedAvg.
Another method trains only on specific portions of local data, which have lower entropy than the average data entropy of all clients. Although this approach reduces overall training time, the performance of the global model cannot surpass FedAvg \cite{bibliography:entropy-to-mitigate-non-iid-data-problem-on-federated-learning-for-the-edge-intelligence-environment}.
The approach proposed by \cite{bibliography:fedentropy-efficient-device-grouping-for-federated-learning-using-maximum-entropy-judgment} employs a refined selection scheme in which clients are first randomly chosen and then re-selected for aggregation based on their soft labels, requiring additional local computation to generate these labels and extra communication to transmit them. Additionally, the filtering process discards already trained client models, wasting their potential contribution to the overall global model.
In contrast, FedEntOpt exchanges label information once before training, keeping communication overhead minimal. Based on this information, it selects clients whose combined label distribution attains maximum entropy, using every trained model during the update process. This leads to more balanced and comprehensive contributions from all classes when aggregating model parameter updates,  improving the performance of the global model compared to state-of-the-art methods.

\section{Problem Formulation}
\Ac{fl} is a distributed learning framework involving multiple clients to collaboratively train a global model without sharing their raw data. We consider \( K \) clients, where each client $k$ has a local dataset \( D^{(k)} = \{(x_i^{(k)}, y_i^{(k)})\}_{i=1}^{n_k} \) of size \( n_k \), where \( x_i^{(k)} \in \mathcal{X} \subseteq \mathbb{R}^d \) and \( y_i^{(k)} \in \mathcal{Y} \subseteq \mathbb{R}^C \) represent the features and labels, respectively. The goal is to learn a global model for image classification \( f_\theta: \mathcal{X} \to \mathcal{Y} \), which is a neural network with parameters \( \theta \in \mathbb{R}^p \), minimizing the following objective:
\begin{align}
\theta^* = \argmin_{\theta \in \mathbb{R}^p } \sum_{k=1}^K \frac{n_k}{\sum_{j=1}^K n_j} \int_{\mathcal{X} \times \mathcal{Y}} \ell(f_\theta(x), y) \, dP_{X^{(k)}, Y^{(k)}},
\label{federated-objective}
\end{align}
where  \(\ell: \mathbb{R}^C \times \mathbb{R}^C \to [0, \infty)\ \) denotes the loss function and \(P_{X^{(k)}, Y^{(k)}} \) the unknown joint probability distribution of each client. It can be expressed as
\begin{align*}
    P_{X^{(k)}, Y^{(k)}}(x, y) = P_{X^{(k)} \mid Y^{(k)} = y}(x) P_{Y^{(k)}}(y).
\end{align*}
The label distribution skew is characterized by the condition that the marginals \( P_{Y^{(k)}} \) vary among the clients, while keeping the class-conditionals  \( P_{X^{(k)} \mid Y^{(k)}} \) identical for all clients. Since these distributions are unknown, the solution of \eqref{federated-objective} is obtained through empirical risk minimization, where the optimization is carried out on a subset of clients \( S \subset A:=\{1,\dots, K\}\):
 \begin{align}
    \Tilde{\theta} =   \sum_{s\in S} \frac{n_s}{\sum_{j \in S} n_j} \underbrace{\argmin_{\theta \in \mathbb{R}^p } \frac{1}{n_s }\sum_{i=1}^{n_s} \ell\left(f_\theta(x_i^{(s)}), y_i^{(s)}\right)}_{= \theta^{(s)}}.
    \label{emprical-ojective}
\end{align}
After receiving the global model from the server, the selected clients update their local models with their data and send the updated model parameters \(\theta^{(s)} \) back for aggregation. This procedure is iterated until convergence \cite{bibliography:communication-efficient-learning-of-deep-networks-from-decentralized-data}. Optimally, the cohort of clients is chosen such that their aggregated local joint distributions approximate the global joint distribution \(P_{X,Y}\), which is the convex combination of the local joints of all  participating clients in the training process according to \eqref{federated-objective}. In this setting, we adopt the common assumption that the global marginal label distribution \(P_{Y}\) is uniform \cite{bibliography:tackling-the-objective-inconsistency-problem-in-heterogeneous-federated-optimization, bibliography:scaffold-stochastic-controlled-averaging-for-federated-learning, bibliography:fedrs-federated-learning-with-restricted-softmax-for-label-distribution-non-iid-data}. Simply averaging model parameters in \eqref{emprical-ojective}  of randomly selected clients can produce a poorly performing global model, since the aggregated local joints of the selected clients,
\begin{align}
    \sum_{s \in S} \frac{n_s}{\sum_{j \in S} n_j} P_{X^{(s)}, Y^{(s)}},
    \label{eq:fl-objective}
\end{align}
can differ significantly from the global distribution \(P_{X,Y}\). FedEntOpt addresses this problem by selecting a subset of clients such that the accumulated label distribution over the selected subset is close to the global label distribution by maximizing its Shannon entropy:
\begin{align}
    P_{Y}(y) = \sum_{s \in S} \frac{n_s}{\sum_{j \in S} n_j} P_{Y^{(s)}}(y).
    \label{eq:label_prob}
\end{align}
Since the local conditionals \(P_{X^{(k)} \mid Y^{(k)}}\) are assumed to be the same for all clients and the global joint distribution is a convex combination of the local ones, approximating the global label distribution by the client subset effectively approximates the global joint distribution.
This in turn reduces the upper bound on excess risk between the empirical federated model by equation \eqref{emprical-ojective} and the optimal centralized model  \cite{bibliography:federated-learning-for-non-iid-data-from-theory-to-algorithm}.

\section{Proposed Method}
To illustrate the proposed method, FedEntOpt, for solving the classification problem in \eqref{federated-objective}, let the label set be defined as \(\mathcal{L} \coloneqq \{1, \dots, C\}\), and the set of available clients \(A \coloneqq\{1, \dots, K\}\). 
Each label \(c  \in \mathcal{L}\) is represented as a one-hot encoded vector \(e_c \in \mathbb{R}^C\). For a client \(k\), we count the number of occurrences of each label and store these counts as a vector \(l^{(k)} \in \mathbb{N}^C\). Specifically, the \(c\)-th component \(l^{(k)}_c\) is given by:
\begin{align*}
        l^{(k)}_c = \sum_{i=1}^{n_k} \mathbf{1}_{\{y_i^{(k)} = e_c\}} \quad \forall c \in \mathcal{L} \ \ \text{and} \ \ \forall k \in A,
\end{align*}
where  \(\mathbf{1}_{\{y_i^{(k)} = e_c\}}\) is an indicator function defined as:
\begin{align*}
    \mathbf{1}_{\{y_i^{(k)} = e_c\}} =
    \begin{cases}
        1 & \text{if } y_i^{(k)} = e_c, \\
        0 & \text{otherwise}.
    \end{cases}
 \end{align*}
Each client sends its label count vector $l^{(k)}$ to the central server once before training, which incurs minimal communication overhead compared to exchanging model parameters during training. A quantitative breakdown of communication costs is provided in the experiment section.  In order to keep track of the aggregated label distribution, the server initializes a vector \(L = 0 \in \mathbb{R}^C\), then samples a first client \(k_0 \in A\) uniformly at random and sets \(L = l^{(k_0)}\).  After that, the server iteratively adds clients to the current subset such that the combined label distribution is maximized. This is achieved by solving the following optimization problem:
\begin{align}
    \argmax_{m\in  A \setminus \{k_0\}} \mathbb{H}\left(\frac{L + l^{(m)}}{\|L + l^{(m)}\|_1}\right),
    \label{fedentopt-objective}
\end{align}
where 
\begin{align*}
  \mathbb{H}(p) := - \sum_{c=1}^C p_c \log_2 (p_c),  
\end{align*}
is the Shannon entropy of the discrete distribution \(p=(p_1,\dots,p_C)\). The label vector associated with the client solving \eqref{fedentopt-objective} is then added to $L$, and the client is excluded from further selection. The process is continued until the desired number of clients is reached. Intuitively, iteratively maximizing entropy encourages adding clients whose data complement missing labels. Since the selection process is deterministic apart from sampling the first client, we maintain a buffer \(B\) of size \(Q\) to exclude recently selected clients and thus prevent repeated selection of the same subset. Whenever a client is chosen, it is placed in \(B\) and excluded from \(A\) until \(B\) reaches its capacity \(Q\). At that point, the clients that were added the earliest are removed, making them available for selection again. To ensure that at least one feasible subset of size $M$ remains selectable, the buffer size must satisfy 
$0 < Q \le K - M$. Algorithm \ref{algorithm:fedentopt} summarizes the proposed approach.

It is worth pointing out that FedEntOpt naturally employs weighting by data size, a strategy commonly adopted by many state-of-the-art federated methods. To illustrate this, pick any \(c \in \mathcal{L} \) and consider the \(c\)-th component, denoted as $L_c$, of the aggregated label vector \(L\) for a selected subset $S$. One obtains:
\begin{align*}
  \frac{L_c}{\|L\|_1}
  &= \frac{\sum_{k \in S} l_c^{(k)}}{\sum_{j \in S}\sum_{r=1}^C l_r^{(j)}}
   = \frac{\sum_{k \in S} n_k \Bigl(\tfrac{1}{n_k}\sum_{i=1}^{n_k} \mathbbm{1}_{\{y_i^{(k)} = e_c\}}\Bigr)}{\sum_{j \in S} n_j} \\
  &= \frac{\sum_{k \in S} n_k \,\tilde{P}_Y^{(k)}(c)}{\sum_{j \in S} n_j}
   = \sum_{k\in S}\Bigl(\frac{n_k}{\sum_{j\in S}n_j}\Bigr)\,\tilde{P}_Y^{(k)}(c),
\end{align*} 
where \(\tilde{P}_Y^{(k)}\) denotes the empirical distribution of labels on client \(k\), serving as an approximation of \(P_Y^{(k)}\).
This expression shows that the aggregated label distribution is exactly a data size weighted mixture of the local distributions.
Consequently, by iteratively maximizing \eqref{fedentopt-objective} to select additional clients, FedEntOpt acts as a greedy algorithm that incrementally increases the entropy of the data size weighted label distribution, effectively approximating the uniform global marginal in \eqref{eq:label_prob}.
\begin{algorithm}[ht]
  \caption{FedEntOpt}
  \label{algorithm:fedentopt}
  \DontPrintSemicolon
  \SetNoFillComment
  \SetKwInOut{Input}{Input}
  \SetKwInOut{Output}{Output}

 \Input{
        \begin{itemize}
            \item \(l^{(k)} = \left[l_1^{(k)}, \dots, l_C^{(k)} \right]\): Label vectors for each client \(k\)
            \item $K$: Total number of clients
            \item $M$: Number of clients to select for the subset
            \item $B = (b_1, b_2, \dots, b_{|B|})$: FIFO buffer (ordered list)
            \item $Q$: Maximum size of the FIFO buffer $B$ assuming  $0 < Q \le K-M$
        \end{itemize}
    }
  \Output{Subset of clients $S$, updated FIFO buffer $B$}

  \(L \leftarrow \mathbf{0} \in \mathbb{R}^C\)\\
  \(S \leftarrow \emptyset\)\\
  \(A \leftarrow \{1,\dots,K\}\) \;
  \For{\(i = 1\) \KwTo \(M\)}{
    \(U \leftarrow A \setminus B\) \;
    \uIf{\(i = 1\)}{
      \(m^* \leftarrow\) Uniform sample from \(U\)
    }
    \Else{
    \(m^* \leftarrow \argmax\limits_{m\in U}\ \mathbb{H}\!\left(\dfrac{L + l^{(m)}}{\|L + l^{(m)}\|_1}\right)\)\;
     
    }
    Append \(m^*\) to \(B\) \;
    \While{$|B| > Q$}{ Remove the oldest client from \(B\) } 
    \(L \leftarrow L + l^{(m^*)}\) \;
    \(S \leftarrow S \cup \{\,m^*\}\) \;  
    \(A \leftarrow A \setminus \{\,m^*\}\) \;  
  }
  \KwRet \(S,\,B\).
\end{algorithm}

The highest possible value for the objective in \eqref{fedentopt-objective} is obtained for a uniform distribution, which is challenging to achieve in practice due to the highly imbalanced data distribution of individual clients. Therefore, the iterative approach for solving \eqref{fedentopt-objective} seeks to come as close to this, which is verified in a short experiment, showing that the selection strategy consistently exhibits high entropy over the course of training. 
We utilize the CIFAR-10 dataset, where we distribute its training data such that each client holds samples from two labels. In each of 100 communication rounds, we select 10 clients to train. For details on the dataset, partitioning strategy, and hyperparameter, refer to the experiment section. In each round, we compute the entropy of the combined label distribution over the selected clients using our proposed method.
We compare our method against pure random selection and the refined approach of Ling et al.\cite{bibliography:fedentropy-efficient-device-grouping-for-federated-learning-using-maximum-entropy-judgment}, which re-evaluates clients based on soft labels after an initial random draw.
The results are visualized in Fig. \ref{fig:entropy-over-rounds}.
\begin{figure}[ht]
    \centering
    \includegraphics[width=\linewidth]{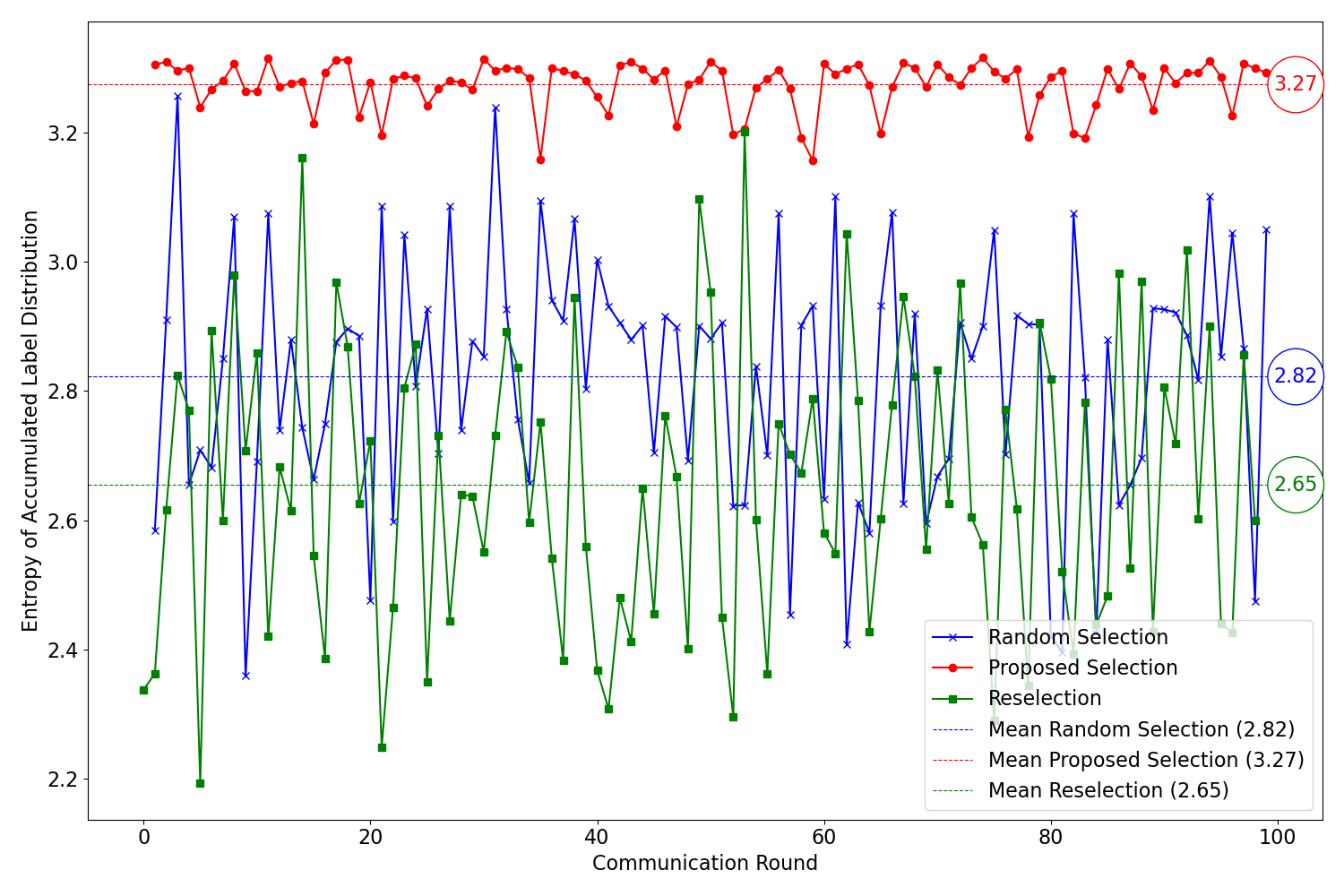}
    \caption{Entropy of the combined label distribution over selected client subsets in each communication round.}
    \label{fig:entropy-over-rounds}
\end{figure}
FedEntOpt consistently achieves the highest entropy with respect to the aggregated label distribution of the selected subset compared to both random selection and the re-selection method, with low variance throughout training.
Interestingly, our results show how random selection tends to have a higher average entropy compared to the re-selection approach. This indicates that decisions based on soft labels can be misleading and lead
to less effective aggregated updates lacking contributions of certain labels due to the lower overall entropy value. One possible reason is that the parameters of the global model might significantly deviate from the optimal parameters for the local data distributions. This divergence means that training on individual clients becomes less effective, since their updates do not align well with the objective of the global model,
making the use of soft labels not informative at all.
Preliminary experiments show that the global model performance of the re-selection method is worse than that of FedAvg due to the lower average entropy. Therefore, we will omit this method in our experiments.
Since the maximum entropy for a $C$-class problem is $\log_2(C)$, an entropy exceeding $\log_2(C-1)$ implies coverage of all labels. Fig. \ref{fig:entropy-over-rounds} shows that for $C=10$, FedEntOpt consistently exceeds $\log_2(9)\approx3.17$. This indicates that despite each client having only two labels, the selected subset collectively covers all ten. 

Although the transmission of the label information does not reveal the raw data, there is still the possibility to reconstruct certain aspects of the data, such as statistical properties. To mitigate this, we will apply differential privacy, which adds noise to the label vector of each client to protect the uploaded label counts, as shown in the experiment section.

\section{Experiments}
To demonstrate the effectiveness of our approach, we implemented FedEntOpt and other baseline algorithms for comparison in PyTorch and ran the associated experiments on a GPU cluster with AMD EPYC 3.5 GHz CPUs, 512 GB RAM, and 4  NVIDIA Ampere A100 GPUs with 40 GB of memory per GPU. Under the label skew setting, FedEntOpt consistently outperforms the other state-of-the-art algorithms in terms of classification accuracy.

\subsection{Experiment Setup}
\subsubsection{Datasets and Preprocessing}
\label{datasets}
For experimental evaluation, we adopt a variety of datasets that include both standard low-resolution vision benchmarks and higher-resolution, domain-specific medical images.
For the low-resolution setting, we use CIFAR-10, CIFAR-100 \cite{bibliography:learning-multiple-layers-of-features-from-tiny-images} and CINIC-10 \cite{bibliography:cinic-10-is-not-imagenet-or-cifar-10}. CIFAR-100 is evaluated in its 20 super-class variant, doubling the number of classes relative to CIFAR-10 (10 classes), while CINIC-10 (10 classes), as an extension to CIFAR-10 with ImageNet images, is split across 200 clients to simulate a larger cohort. 
To test on higher-resolution, real-world images, we adopt the RGB versions of PathMNIST (9 classes) and TissueMNIST (8 classes) from MedMNIST \cite{bibliography:medmnist}, together with the Brain-Tumor MRI data (4 classes) released on Kaggle \cite{bibliography:brain-tumor}. While CIFAR-style images are resized to 32 × 32, for medical images, the size 224 × 224 is used. All inputs are scaled to $[0,1]$, standardized per channel, and augmented with color jitter, horizontal flip ($p=0.5$) and random grayscale ($p=0.1$).  The training splits are distributed among clients, while the uniformly distributed validation splits are used to evaluate the classification accuracy of the global model after aggregation.

\subsubsection{Simulation of Label Skew}
\label{sec:label-skew}
To simulate label skew, we adopt both quantity- and distribution-based partitioning strategies as outlined in \cite{bibliography:federated-learning-on-non-iid-data-silos}. In the quantity-based case, each client holds samples of only \(j\) classes with \(j < C\). We denote this by $C = j$  to indicate that a client has data for only $j$ distinct classes.
In the distribution-based case, we allocate samples according to \(p_j\sim\mathrm{Dir}_K(\beta)\), where \( p_j \) is a \( K \)-dimensional vector representing the proportions of instances of class \( j \) assigned to each client. Specifically, client \(k\) receives \(p_{j,k}\) samples of class \(j\). For simplicity, we refer to this partitioning strategy as \(\mathrm{Dir}(\beta)\).

\subsubsection{Model Architecture}
Since our experiments focus on image classification, we employ different convolutional neural network (CNN) architectures.
For CIFAR-style datasets, clients train a lightweight LeNet-5 model \cite{bibliography:gradient-based-learning-applied-to-document-recognition} or a larger VGG-11 model  \cite{bibliography:vgg11}, reflecting heterogeneous hardware capacities. For the medical datasets, we follow a similar strategy and employ EfficientNet-B0 and the larger EfficientNet-B3 \cite{bibliography:efficientnet}, which outperform VGG and other common CNN architectures in this domain \cite{bibliography:medical-models}.

\subsubsection{Baseline Methods}
We compare FedEntOpt with two sets of baselines. The first group randomly selects clients and handles label imbalance locally, inducing FedAvg \cite{bibliography:communication-efficient-learning-of-deep-networks-from-decentralized-data}, FedProx \cite{bibliography:federated-optimization-in-heterogeneous-networks},  FedNova \cite{bibliography:tackling-the-objective-inconsistency-problem-in-heterogeneous-federated-optimization}, SCAFFOLD \cite{bibliography:scaffold-stochastic-controlled-averaging-for-federated-learning}, FedRS \cite{bibliography:fedrs-federated-learning-with-restricted-softmax-for-label-distribution-non-iid-data} 
and FedLC \cite{bibliography:federated-learning-with-label-distribution-skew-via-logits-calibration}. The second group addresses label skew at the server by requiring clients to transmit label information, including FedConcat \cite{bibliography:exploiting-label-skews-in-federated-learning-with-model-concatenation} and the method from \cite{bibliography:addressing-class-imbalance-in-federated-learning}. Since it has no specific name, we refer to it as KL because it relies on KL-divergence to select a subset of clients for training.

\subsubsection{Hyperparameter and Simulation Settings}
All models are trained locally for five epochs with the SGD optimizer using a momentum of 0.9. The local batch size is set to 64. The initial learning rate is \(0.01\) for the CIFAR-style datasets and \(0.001 \) for the  medical images, and decays by 0.98 after every communication round. 
We use cross-entropy loss and tune the weight decay from \(\{0.0005,\,0.001,\,0.0025\}\), selecting \(0.0005\) for LeNet-5 and EfficientNet-B0/B3 and \(0.001\) for VGG-11.
We train for 500 communication rounds on CIFAR-style datasets and 200 rounds on medical datasets, evaluating the global model accuracy on the test set after each round. To ensure fairness, we use the standard FedAvg weighting based on client data size to aggregate updates, as employed by all baselines.
We set the available client pool to 100 clients for CIFAR-10/100 and MedMNIST,  200 clients for CINIC-10 due to its larger training set, and 40 clients for the smaller Brain-tumor MRI data. In each round, we select 10\% of the clients for CIFAR-style datasets and 5\% for medical datasets, reflecting realistic participation rates and challenging learning conditions. Unless stated otherwise, we follow this configuration.  For FedEntOpt, we use the best-performing buffer size
$Q \in \{ 50\%, 60\%, 70\%, 80\%\}$ of the available clients. We find that $Q = 50\%$ performs the best for distribution-based partitioning, while $Q = 70\%$ is optimal for quantity-based partitioning. The hyperparameters of the baselines follow the original papers: FedRS uses $\alpha = 0.5$, FedLC uses $\tau = 1$, and FedConcat employs five clusters with the same total communication budget as FedAvg. For FedProx, we tune the proximal parameter $\mu \in \{0.0001, 0.001, 0.01\}$. For both partitioning strategies, the optimal value is found to be \(\mu = 0.0001\). Each experiment is run on 3 random seeds. The mean and standard deviation of the averaged results from the last 10 rounds are then reported, providing a fair evaluation \cite{bibliography:experiment-evaluation}. Experiments cover the two most challenging label-skew settings: distribution-based imbalances with \( \mathrm{Dir}(0.1) \) and quantity-based imbalances with \( C = 2 \). Note that \( \mathrm{Dir}(0.1) \) is considered the most challenging scenario because samples of certain classes are heavily concentrated among a few clients.

\subsection{Experiment Results for CIFAR-Style Data}
\subsubsection{Baseline Comparison}
On the low-resolution CIFAR-style datasets, we evaluate FedEntOpt against baseline methods using the LeNet-5 and VGG-11 architectures, simulating clients with heterogeneous hardware capabilities. Our method consistently achieves higher classification accuracy across all datasets and partitions, as illustrated in Table \ref{tab:vision_combined}. 
\begin{table*}[htbp]
    \centering
    \caption{Mean test accuracy (\%) over final 10 rounds on CIFAR-10, CINIC-10, and CIFAR-100 with
LeNet-5 and VGG-11 architecture, comparing the baselines with FedEntOpt. Bold values indicate the best and second-best performance among the compared methods.}
    \setlength{\tabcolsep}{1mm}
    \resizebox{\textwidth}{!}{
     \renewcommand{\arraystretch}{1.4}
    \begin{tabular}{|c|c|c|c|c|c|c|c|c|c|c|c|c|}
    \hline
    \textbf{Model} & \textbf{Dataset} & \textbf{Partition} & \textbf{FedAvg} & \textbf{FedProx} & \textbf{FedNova} & \textbf{SCAFFOLD} & \textbf{FedConcat} & \textbf{FedLC} & \textbf{FedRS} & \textbf{KL} & \textbf{FedEntOpt} & \textbf{FedEntOpt (DP)} \\ 
    \hline

    \multirow{6}{*}{LeNet-5}
    & \multirow{2}{*}{CIFAR-10} & $\mathrm{Dir}(0.1)$  
        & $46.61 \pm 0.17$ & $44.84 \pm 0.20$ & $10.00 \pm 0.00$ & $14.12 \pm 0.00$ & $23.85 \pm 3.17$ 
        & $40.50 \pm 0.05$ & $38.45 \pm 0.07$ & $34.76 \pm 0.01$ 
        & $\mathbf{52.80 \pm 0.03}$ & $\mathbf{52.29 \pm 0.07}$ \\ 
    \cline{3-13}
    &                               & $C = 2$      
        & $51.58 \pm 0.07$ & $51.68 \pm 0.09$ & $51.71 \pm 0.03$ & $32.21 \pm 0.01$ & $24.45 \pm 3.32$ 
        & $49.83 \pm 0.01$ & $49.40 \pm 0.01$ & $33.06 \pm 0.00$ 
        & $\mathbf{53.84 \pm 0.03}$ & $\mathbf{53.18 \pm 0.04}$ \\ 
    \cline{2-13}

    & \multirow{2}{*}{CINIC-10}    & $\mathrm{Dir}(0.1)$  
        & $37.84 \pm 0.19$ & $37.71 \pm 0.22$ & $10.00 \pm 0.00$ & $10.04 \pm 0.00$ & $20.51 \pm 2.94$ 
        & $29.44 \pm 0.02$ & $29.34 \pm 0.03$ & $27.41 \pm 0.01$ 
        & $\mathbf{40.11 \pm 0.04}$ & $\mathbf{40.10 \pm 0.02}$ \\ 
    \cline{3-13}
    &                               & $C = 2$      
        & $40.89 \pm 0.05$ & $40.90 \pm 0.06$ & $39.24 \pm 0.06$ & $14.05 \pm 0.01$ & $18.05 \pm 2.14$ 
        & $38.17 \pm 0.01$ & $39.28 \pm 0.01$ & $26.54 \pm 0.02$ 
        & $\mathbf{41.41 \pm 0.04}$ & $\mathbf{42.06 \pm 0.01}$ \\ 
    \cline{2-13}

    & \multirow{2}{*}{CIFAR-100}   & $\mathrm{Dir}(0.1)$  
        & $31.06 \pm 0.19$ & $29.87 \pm 0.35$ & $30.97 \pm 0.12$ & $21.52 \pm 0.08$ & $12.91 \pm 1.81$ 
        & $27.00 \pm 0.05$ & $25.73 \pm 0.98$ & $18.32 \pm 0.01$ 
        & $\mathbf{32.71 \pm 0.11}$ & $\mathbf{32.82 \pm 0.20}$ \\ 
    \cline{3-13}
    &                               & $C = 2$      
        & $24.40 \pm 0.09$ & $25.61 \pm 0.13$ & $24.66 \pm 0.10$ & $ \hspace{1.3mm}7.44 \pm 0.01$  & $\hspace{1.3mm} 6.68 \pm 0.92$  
        & $24.63 \pm 0.07$ & $25.83 \pm 0.16$ & $17.89 \pm 0.02$ 
        & $\mathbf{27.57 \pm 0.13}$ & $\mathbf{27.44 \pm 0.11}$ \\ 
    \hline

    \multirow{6}{*}{VGG-11}
    & \multirow{2}{*}{CIFAR-10}    & $\mathrm{Dir}(0.1)$  
        & $58.33 \pm 0.05$ & $58.45 \pm 0.07$ & $49.66 \pm 0.09$ & $61.89 \pm 0.08$ & $10.00 \pm 0.00$ 
        & $54.27 \pm 0.08$ & $53.87 \pm 0.05$ & $39.98 \pm 0.02$ 
        & $\mathbf{63.79 \pm 0.09}$ & $\mathbf{64.17 \pm 0.03}$ \\ 
    \cline{3-13}
    &                               & $C = 2$      
        & $64.32 \pm 0.06$ & $64.62 \pm 0.03$ & $64.47 \pm 0.05$ & $60.08 \pm 0.07$ & $10.00 \pm 0.00$ 
        & $54.28 \pm 0.04$ & $55.52 \pm 0.07$ & $31.56 \pm 0.03$ 
        & $\mathbf{65.26 \pm 0.01}$ & $\mathbf{66.10 \pm 0.02}$ \\ 
    \cline{2-13}

    & \multirow{2}{*}{CINIC-10}    & $\mathrm{Dir}(0.1)$  
        & $43.77 \pm 0.08$ & $43.66 \pm 0.06$ & $42.36 \pm 0.04$ & $46.09 \pm 0.02$ & $10.00 \pm 0.00$ 
        & $30.76 \pm 0.06$ & $41.24 \pm 0.05$ & $25.81 \pm 0.08$ 
        & $\mathbf{47.01 \pm 0.02}$ & $\mathbf{47.71 \pm 0.01}$ \\ 
    \cline{3-13}
    &                               & $C = 2$      
        & $46.82 \pm 0.07$ & $46.84 \pm 0.03$ & $\mathbf{47.11 \pm 0.08}$ & $44.67 \pm 0.01$ & $10.00 \pm 0.00$ 
        & $31.85 \pm 0.01$ & $45.14 \pm 0.03$ & $24.87 \pm 0.04$ 
        & $46.26 \pm 0.07$ & $\mathbf{47.21 \pm 0.08}$ \\ 
    \cline{2-13}

    & \multirow{2}{*}{CIFAR-100}   & $\mathrm{Dir}(0.1)$  
        & $39.96 \pm 0.02$ & $39.94 \pm 0.01$ & $39.73 \pm 0.05$ & $30.67 \pm 0.04$ & $5.00 \pm 0.00$ 
        & $37.30 \pm 0.07$ & $37.92 \pm 0.06$ & $25.37 \pm 0.05$ 
        & $\mathbf{42.80 \pm 0.08}$ & $\mathbf{43.00 \pm 0.04}$ \\ 
    \cline{3-13}
    &                               & $C = 2$      
        & $31.53 \pm 0.09$ & $31.36 \pm 0.08$ & $30.51 \pm 0.03$ & $29.77 \pm 0.06$ & $5.00 \pm 0.00$ 
        & $17.77 \pm 0.03$ & $24.24 \pm 0.09$ & $21.34 \pm 0.01$ 
        & $\mathbf{33.55 \pm 0.02}$ & $\mathbf{33.29 \pm 0.06}$ \\ 
    \hline
    \end{tabular}
    }
    \label{tab:vision_combined}
\end{table*}
For the LeNet-5 architecture, the most gain, exceeding 6\% compared to the second-best algorithm, is observed on the CIFAR-10 dataset for \(\mathrm{Dir}(0.1)\). More moderate but still consistent gains of around 1–3 percentage points are seen on CINIC-10 and CIFAR-100. We also noted that FedConcat consistently underperformed, suggesting that, under partial participation, the learned cluster representations are insufficient. This can be explained by the random client selection, which results in some clusters being trained less frequently than others.
To address this issue, the authors use a participation rate of 50\%, which is impractical in real-world federated learning scenarios, where client availability is limited.
Additionally, KL converges at a low accuracy, suggesting overfitting to a subset of clients, as indicated in the related work section. FedNova and  SCAFFOLD, as noted by \cite{bibliography:federated-learning-on-non-iid-data-silos}, suffer from low classification accuracy, especially in the case of \(\mathrm{Dir}(0.1)\). 
A similar pattern is observed with the VGG-11 architecture.
On CIFAR-100 under \(\mathrm{Dir}(0.1)\), FedEntOpt outperforms the best baseline by about 3 percentage points, while on the other datasets it exceeds the best baseline by approximately 1–2 percentage points. Consistent with the LeNet-5 results, FedConcat again underperforms during training, and this degradation is even more pronounced for the larger VGG-11 architecture. Its increased complexity and parameter count make it more sensitive to imbalances in training. In contrast to the LeNet-5 architecture, FedNova and SCAFFOLD perform significantly better.
\label{fig:dir-0.1-lenet5}
For both architectures, FedEntOpt not only maintains superior accuracy but also reaches a given target accuracy significantly faster than competing methods.

\subsubsection{Differential Privacy}
\label{sec:differential-privacy}
Since FedEntOpt instructs each client to reveal its label information, we refer to the principles of differential privacy to have a mathematical guarantee for data protection.
It is defined according to \cite{bibliography:the-algorithmic-foundations-of-differential-privacy}, which is adapted to our setting:
\begin{definition}
A randomized algorithm \( \mathcal{A} \) with domain \( \mathbb{N}^C \) is \( \epsilon \)-differentially private if for any two label count vectors \( d, d' \in \mathbb{N}^C \)  with \( \|d-d'\|_1=1\), and for any subset of possible outputs \( S \subseteq \text{Range}(\mathcal{A}) \), the following inequality holds:
\begin{align}
    \Pr[\mathcal{A}(d) \in S] \leq e^{\epsilon} \Pr[\mathcal{A}(d') \in S],
    \label{eq:privacy}
\end{align}
where the probability is taken over the randomness used by the algorithm.
\end{definition}
This definition ensures that the presence or absence of a single data point in the label count vector does not significantly change the probability distribution over the output of the algorithm $\mathcal{A}$.
To mathematically ensure that \eqref{eq:privacy} holds, the Laplace mechanism is used, which adds noise to each component of the label count vector \cite{bibliography:the-algorithmic-foundations-of-differential-privacy}. Specifically, for client \(k\), we obtain:
\begin{align}
    \label{eq:laplace}
    \mathcal{A}(l^{(k)}) = l^{(k)} + (Z_1, Z_2, \ldots, Z_C),
\end{align}
where each \( Z_i \sim \text{Lap}\left(\frac{1}{\epsilon} \right)  \).
Smaller values of \( \epsilon \) provide stronger privacy, as more noise is added to each component of $l^{(k)}$. To test the robustness of FedEntOpt using the Laplace mechanism, we choose a privacy budget of $\epsilon = 0.5$, which provides a strong level of data protection. According to \eqref{eq:laplace}, we apply this mechanism once to the label count vector of each client without any further preprocessing.
The modified vectors are then uploaded to the central server before the first federated learning round begins. The global model performance in this setup is shown in the last column of Table \ref{tab:vision_combined}, denoted as FedEntOpt (DP). It can be seen that FedEntOpt manages to maintain the same high classification accuracy, matching the performance of the non-private variant. Interestingly, in some cases the performance is even increased. The findings demonstrate that the proposed method outperforms several baseline algorithms even under strong privacy guarantees regardless of the model architecture.

\subsubsection{Impact of Client Participation Rate}
\label{low-client-participation-rate}
To test the robustness of FedEntOpt with different subset sizes of clients, we vary the baseline participation rate of $10\%$ and investigate the performance of all methods in the range of $4\%$ to $9\%$. This effectively simulates real-world scenarios where clients are not always available for the training process. 
We demonstrate this using the partitioning strategies $C=2$ and \(\mathrm{Dir}(0.1)\)  for both CIFAR-10 and CIFAR-100.
For $C=2$ on CIFAR-100, the results are excluded as they are nearly identical across all methods. This can be attributed to the relative simplicity of this partitioning strategy compared to \(\mathrm{Dir}(0.1)\).
Figure \ref{fig:combined-varying-client-size-fraction}  illustrates that FedEntOpt consistently maintains high accuracy, even with lower participation rates regardless of the model size, while the other methods partially suffer performance degradation. Moreover, because FedEntOpt achieves comparable accuracy with fewer participating clients, it also reduces overall communication overhead. The difference in classification accuracy is particularly pronounced in the challenging $\mathrm{Dir}(0.1)$ case, making FedEntOpt the default choice to effectively handle both label skew and low participation scenarios.
\begin{figure*}[ht]
\centering
\includegraphics[width=\textwidth]{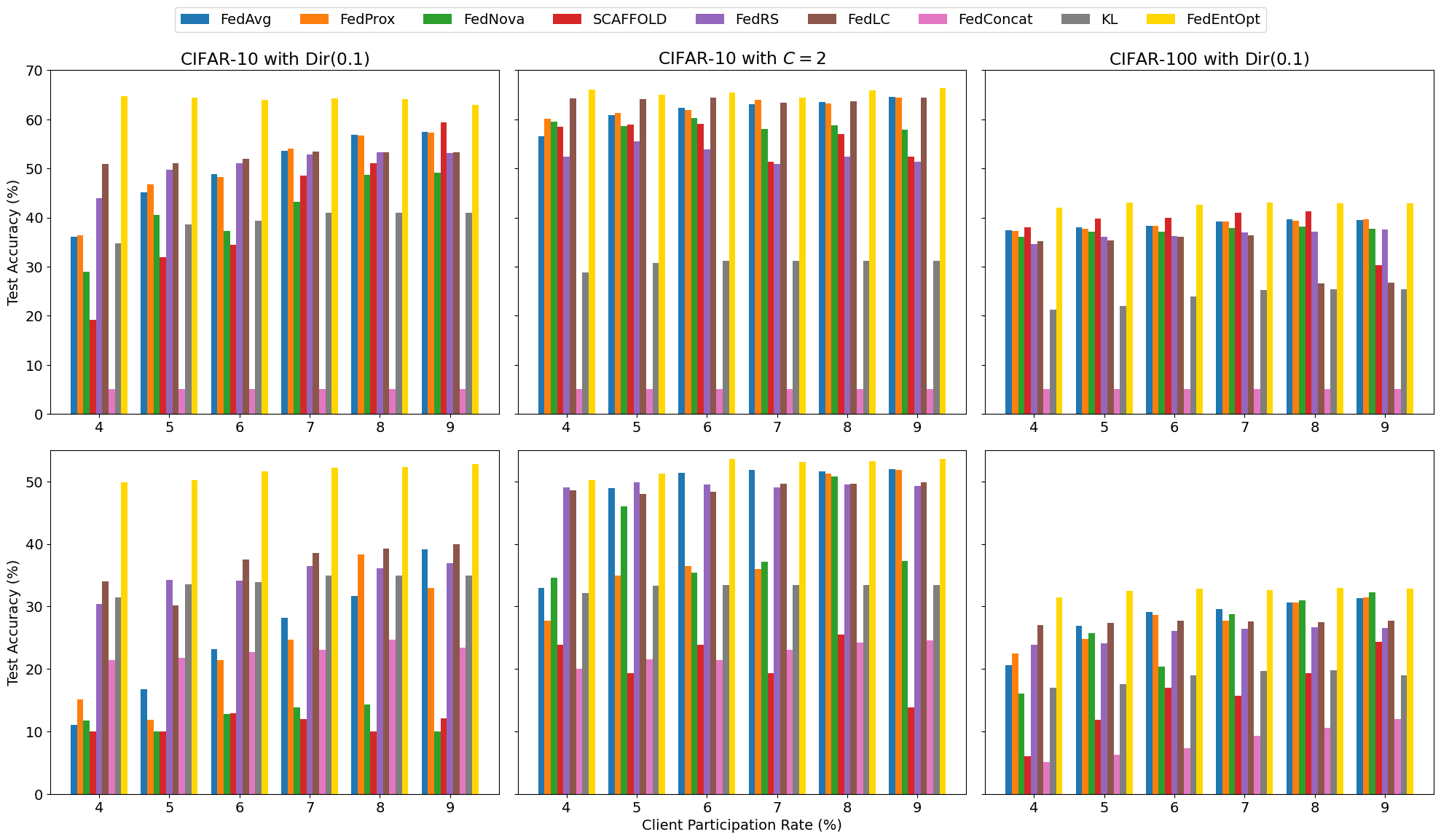}
\caption{Mean test accuracy (\%) over 10 final rounds of varying participation rates for $\mathrm{Dir}(0.1)$ and $C=2$ on CIFAR-10 and CIFAR-100, comparing baselines with FedEntOpt. Results for VGG-11 are shown above and for LeNet-5 below.}
\label{fig:combined-varying-client-size-fraction}
\end{figure*}

\subsubsection{Combination of FedEntOpt with other Baselines}
Since the client selection strategy is orthogonal to methods like FedProx, FedNova, SCAFFOLD, FedRS, and FedLC, which choose clients randomly and handle heterogeneity on the client side, we demonstrate that combining these methods with FedEntOpt can enhance their performance. The results are presented in Table \ref{tab:combined-lenet-vgg}, illustrating the absolute improvement in classification accuracy compared to the baseline performance according to Table \ref{tab:vision_combined}.
Especially in the case of distribution-based imbalances, the classification accuracy increases significantly by more than 40\% for LeNet-5, while VGG-11 also sees substantial gains of over 13\%.  Even under the simpler $C=2$ partition, FedEntOpt still yields modest improvements, although the absolute gains are smaller.
This shows that FedEntOpt can further enhance classification accuracy even for algorithms that already handle heterogeneity on the client side.
\begin{table}[ht]
    \centering
    \caption{Absolute change in mean test accuracy (\%) over final 10 rounds when each baseline is combined with FedEntOpt on CIFAR-10, CINIC-10, and CIFAR-100. In every cell, the first value refers to LeNet-5 architecture and the second to VGG-11 architecture. \greenup\ indicates an absolute increase, while \reddown\ indicates a decrease in classification accuracy, compared to baseline performance.}
    \resizebox{\columnwidth}{!}{%
    \renewcommand{\arraystretch}{1.35}
   \begin{tabular}{|c|c|r|r|r|}
     \hline
    \textbf{Partition} & \textbf{Algorithm} &
    \multicolumn{1}{|c|}{\textbf{CIFAR-10}} &
    \multicolumn{1}{c|}{\textbf{CINIC-10}} &
    \multicolumn{1}{c|}{\textbf{CIFAR-100}} \\ \hline
    \multirow{5}{*}{$\mathrm{Dir}(0.1)$} 
     & FedProx   & 7.67\greenup/4.80\greenup    & 2.42\greenup/3.78\greenup    & 2.48\greenup/2.59\greenup   \\ \cline{2-5}
     & FedNova   & 41.16\greenup/13.28\greenup  & 26.74\greenup/4.62\greenup    & 2.09\greenup/2.67\greenup   \\ \cline{2-5}
     & SCAFFOLD  & 33.66\greenup/4.69\greenup    & 27.82\greenup/1.55\reddown   & 11.82\greenup/13.87\greenup \\ \cline{2-5}
     & FedRS     & 9.21\greenup/3.70\greenup     & 7.85\greenup/5.17\greenup     & 5.35\greenup/1.90\greenup   \\ \cline{2-5}
     & FedLC     & 7.31\greenup/2.80\greenup     & 7.22\greenup/3.50\greenup     & 4.11\greenup/2.09\greenup   \\ \hline
    \multirow{5}{*}{$C = 2$} 
     & FedProx   & 0.85\greenup/1.45\greenup     & 0.42\greenup/0.35\reddown     & 6.74\greenup/1.34\greenup   \\ \cline{2-5}
     & FedNova   & 2.17\greenup/1.06\greenup     & 2.44\greenup/0.60\reddown     & 5.11\greenup/3.11\greenup   \\ \cline{2-5}
     & SCAFFOLD  & 20.16\greenup/1.99\greenup    & 27.02\greenup/2.38\greenup    & 13.95\greenup/6.39\reddown  \\ \cline{2-5}
     & FedRS     & 0.18\greenup/0.80\greenup     & 1.08\greenup/0.50\greenup     & 1.90\greenup/0.27\reddown   \\ \cline{2-5}
     & FedLC     & 0.14\greenup/1.79\reddown     & 0.25\reddown/10.62\greenup    & 0.98\greenup/0.99\reddown   \\ \hline
    \end{tabular}
    }
    \label{tab:combined-lenet-vgg}
\end{table}

\subsubsection{Client Availability Simulation}
We evaluate FedEntOpt under dropout scenarios where clients become unavailable after being selected due to factors such as unstable network connectivity or power constraints. Additionally, we consider the straggling case, where clients cannot complete their local training in a fixed time before the server requests updates for aggregation. 
For simulation, we selected FedAvg and FedProx as baseline algorithms because they demonstrated the most consistent performance across the considered label skew settings (see Table \ref{tab:vision_combined}). To simulate client dropout, we applied a fixed dropout probability of 30\%, which means that after selecting a subset of clients, 30\% became unavailable for local training. In the straggling case, we designated 50\% of the available client pool as stragglers, assigning each a random number of local epochs between one and five that the client was able to train. The results of the experiments are presented in Table \ref{tab:dropout-straggling} for both the LeNet-5 and VGG-11 architectures.
\begin{table}[ht]
    \small
    \centering
    \caption{Mean test accuracy (\%) over final 10 rounds under dropout and straggling  on CIFAR-10 and CIFAR-100, comparing FedAvg and FedProx with FedEntOpt. Bold values indicate the best performance among the compared methods.}
    \setlength{\tabcolsep}{1mm}
     \resizebox{\columnwidth}{!}{%
         \renewcommand{\arraystretch}{1.35}
    \begin{tabular}{|c|c|c|c|c|c|c|}
    \hline
    \textbf{Simulation Type} & \textbf{Dataset} & \textbf{Partition} & \textbf{Model} & \textbf{FedAvg} & \textbf{FedProx} & \textbf{FedEntOpt} \\ 
    \hline
    \multirow{8}{*}{Dropout}
      & \multirow{4}{*}{CIFAR-10}
        & \multirow{2}{*}{$C=2$} 
          & LeNet-5   & $36.85 \pm 0.08$ & $37.27 \pm 0.04$ & $\mathbf{52.36 \pm 0.06}$ \\ \cline{4-7}
        &  
        &  
          & VGG-11    & $54.08 \pm 0.05$ & $54.60 \pm 0.03$ & $\mathbf{62.74 \pm 0.07}$ \\ \cline{3-7}
      &  
        & \multirow{2}{*}{$\mathrm{Dir}(0.1)$} 
          & LeNet-5   & $11.47 \pm 0.02$ & $17.39 \pm 0.03$ & $\mathbf{51.03 \pm 0.07}$ \\ \cline{4-7}
        &  
        &  
          & VGG-11    & $63.29 \pm 0.02$ & $62.71 \pm 0.06$ & $\mathbf{63.59 \pm 0.04}$ \\ 
    \cline{2-7}
      & \multirow{4}{*}{CIFAR-100}
        & \multirow{2}{*}{$C=2$} 
          & LeNet-5   & $19.73 \pm 0.09$ & $20.97 \pm 0.05$ & $\mathbf{23.29 \pm 0.01}$ \\ \cline{4-7}
        &  
        &  
          & VGG-11    & $38.81 \pm 0.08$ & $38.32 \pm 0.03$ & $\mathbf{42.42 \pm 0.05}$ \\ \cline{3-7}
      &  
        & \multirow{2}{*}{$\mathrm{Dir}(0.1)$} 
          & LeNet-5   & $29.40 \pm 0.06$ & $29.35 \pm 0.07$ & $\mathbf{32.47 \pm 0.02}$ \\ \cline{4-7}
        &  
        &  
          & VGG-11    & $29.42 \pm 0.01$ & $29.28 \pm 0.07$ & $\mathbf{32.29 \pm 0.02}$ \\ 
    \hline

    \multirow{8}{*}{Straggling}
      & \multirow{4}{*}{CIFAR-10}
        & \multirow{2}{*}{$C=2$} 
          & LeNet-5   & $49.70 \pm 0.03$ & $50.45 \pm 0.02$ & $\mathbf{51.71 \pm 0.04}$ \\ \cline{4-7}
        &  
        &  
          & VGG-11    & $56.32 \pm 0.06$ & $56.59 \pm 0.02$ & $\mathbf{62.27 \pm 0.09}$ \\ \cline{3-7}
      &  
        & \multirow{2}{*}{$\mathrm{Dir}(0.1)$} 
          & LeNet-5   & $44.92 \pm 0.08$ & $43.71 \pm 0.05$ & $\mathbf{52.36 \pm 0.03}$ \\ \cline{4-7}
        &  
        &  
          & VGG-11    & $63.20 \pm 0.03$ & $63.80 \pm 0.04$ & $\mathbf{63.80 \pm 0.05}$ \\ 
    \cline{2-7}
      & \multirow{4}{*}{CIFAR-100}
        & \multirow{2}{*}{$C=2$} 
          & LeNet-5   & $24.09 \pm 0.07$ & $24.72 \pm 0.06$ & $\mathbf{26.22 \pm 0.08}$ \\ \cline{4-7}
        &  
        &  
          & VGG-11    & $39.22 \pm 0.02$ & $39.14 \pm 0.05$ & $\mathbf{41.68 \pm 0.03}$ \\ \cline{3-7}
      &  
        & \multirow{2}{*}{$\mathrm{Dir}(0.1)$} 
          & LeNet-5   & $28.33 \pm 0.02$ & $28.43 \pm 0.01$ & $\mathbf{31.30 \pm 0.09}$ \\ \cline{4-7}
        &  
        &  
          & VGG-11    & $31.60 \pm 0.07$ & $31.67 \pm 0.01$ & $\mathbf{32.71 \pm 0.06}$ \\ 
    \hline
    \end{tabular}
    }
    \label{tab:dropout-straggling}
\end{table}
FedEntOpt consistently outperforms FedAvg and FedProx across all tested scenarios, enhancing classification accuracy by over 30\% under dropout and up to  8\% in straggling conditions, considering both model architectures. Interestingly, FedEntOpt does not seem to be affected under this setup, maintaining performance comparable to the baseline case according to Table \ref{tab:vision_combined}.

\subsection{Experiment Results for Medical Data}
Finally, we evaluate the proposed method on the medical data, using a setup analogous to the CIFAR‐style experiments and employing both the smaller EfficientNet-B0 and the larger EfficientNet-B3 architectures. To simulate real‐world conditions, we employ a client participation rate of 5\% and test the most challenging partitioning strategy \(\mathrm{Dir}(0.1)\). Table \ref{tab:medical_full} summarizes the results.
\begin{table}[ht]
\small
\centering
\caption{Mean test accuracy (\%) over final 10 rounds on medical datasets PathMNIST, TissueMNIST, and Brain-Tumor MRI Data on
\(\mathrm{Dir}(0.1)\). EfficientNet-B0 and EfficientNet-B3 are abbreviated as B0 and B3, respectively. Bold values indicate the best performance among the compared methods.}
 \resizebox{\columnwidth}{!}{%
       \renewcommand{\arraystretch}{1.35}
\begin{tabular}{|c|c|c||c|c||c|c|}
\hline
\multirow{2}{*}{\textbf{Algorithm}}
& \multicolumn{2}{c||}{\textbf{PathMNIST}}
& \multicolumn{2}{c||}{\textbf{TissueMNIST}}
& \multicolumn{2}{c|}{\textbf{Brain-Tumor MRI}}\\
\cline{2-7}
& \textbf{B0} & \textbf{B3} & \textbf{B0} & \textbf{B3} & \textbf{B0} & \textbf{B3}\\
\hline
FedAvg     & $82.26 \pm 0.04$ & $82.56 \pm 0.06$ & $36.55 \pm 0.07$ & $40.88 \pm 0.06$ & $60.02 \pm 0.07$ & $58.52 \pm 0.02$ \\ \hline
FedProx    & $82.24 \pm 0.04$ & $82.63 \pm 0.03$ & $37.20 \pm 0.04$ & $41.02 \pm 0.05$ & $60.57 \pm 0.07$ & $56.70 \pm 0.08$ \\ \hline
FedNova    & $50.43 \pm 0.06$ & $53.77 \pm 0.05$ & $41.10 \pm 0.05$ & $34.21 \pm 0.06$ & $57.78 \pm 0.05$ & $56.50 \pm 0.01$ \\ \hline
SCAFFOLD   & $87.59 \pm 0.03$ & $88.52 \pm 0.02$ & $44.64 \pm 0.03$ & $44.38 \pm 0.02$ & $64.21 \pm 0.03$ & $64.54 \pm 0.05$ \\ \hline
FedRS      & $76.56 \pm 0.04$ & $78.73 \pm 0.07$ & $33.97 \pm 0.05$ & $31.54 \pm 0.08$ & $51.55 \pm 0.06$ & $48.30 \pm 0.08$ \\ \hline
FedLC      & $59.93 \pm 0.06$ & $79.83 \pm 0.04$ & $39.81 \pm 0.02$ & $33.65 \pm 0.03$ & $45.44 \pm 0.05$ & $44.47 \pm 0.08$ \\ \hline
KL         & $56.62 \pm 0.02$ & $62.12 \pm 0.06$ & $28.10 \pm 0.04$ & $35.22 \pm 0.09$ & $45.40 \pm 0.02$ & $49.53 \pm 0.02$ \\ \hline
\textbf{FedEntOpt} & $\mathbf{90.17 \pm 0.05}$ & $\mathbf{90.66 \pm 0.02}$ &
$\mathbf{50.67 \pm 0.05}$ & $\mathbf{48.05 \pm 0.07}$ &
$\mathbf{69.65 \pm 0.06}$ & $\mathbf{66.76 \pm 0.03}$ \\ \hline
\end{tabular}
}
\label{tab:medical_full}
\end{table}

Across the three medical datasets and both the EfficientNet-B0 and B3 models, FedEntOpt consistently achieves the highest classification accuracy. Although SCAFFOLD is the second-best performing method, its accuracy is consistently 2–6 percentage points lower across the different datasets. This shows that FedEntOpt also performs well on higher-resolution data, even when the number of participating clients in each round is limited.

\subsection{Additional Experiments}
\subsubsection{Communication Cost of FedEntOpt}
\label{sec:overhead}
FedEntOpt requires that each client $k$ uploads its label count vector $l^{(k)} \in \mathbb{N}^C$ once before training. To handle both scenarios of uploading raw label counts and perturbed label counts under differential privacy, we assume that all transmitted scalars are represented as 32-bit floating-point numbers, so each entry occupies 4 bytes. Thus, sending the label count vector costs $4C$ bytes per client, which amounts to $4CK$ bytes in aggregate across all clients.
Meanwhile, in every communication round, each selected client uploads its model parameters to the server. Assuming that each client model has $p$ parameters, i.e., $|\theta^{(k)}| = p$ for all $k \in \{1,\dots, K \}$ , the upload cost per round is $4p$ bytes. 
For the models utilized in our experiments, sending label counts from all $K$ clients is several orders of magnitude cheaper than uploading model parameters of a single client model, as summarized in Table \ref{tab:comm-totals}.
\begin{table}[ht]
    \small
    \centering
       \caption{Communication cost across all models and datasets, comparing the one-time upload of all clients’ label-count vectors of size \(4CK\) bytes, denoted as Labels,  with the per-round upload of one client’s model parameters of size \(4p\) bytes, referred to as Model. Ratios in the last column are computed from exact byte counts and rounded to 4 significant figures. Sizes are in megabytes (MB \(=10^6\) bytes).}
    \setlength{\tabcolsep}{1mm}
    \resizebox{\columnwidth}{!}{%
        \renewcommand{\arraystretch}{1.35}
        \begin{tabular}{|c|c|c|c|c|c|}
        \hline
        \textbf{Dataset} & \textbf{$C$} & \textbf{$K$} & \textbf{Labels (MB)} & \textbf{Model (MB)} & \textbf{Model / Labels} \\
        \hline
        \multirow{2}{*}{CIFAR-10} & \multirow{2}{*}{10} & \multirow{2}{*}{100}
            & \multirow{2}{*}{$0.0040$}
            & \hspace{2.1em} LeNet-5: $0.247$ & \hspace{2em} $61.75\times$ \\
        \cline{5-6}
        & & & & \hspace{2.4em} VGG-11: $36.800$ & $9.200{\times}10^{3}\times$ \\
        \hline
        \multirow{2}{*}{CINIC-10} & \multirow{2}{*}{10} & \multirow{2}{*}{200}
            & \multirow{2}{*}{$0.0080$}
            & \hspace{2.1em} LeNet-5: $0.247$ & \hspace{2em} $30.88\times$ \\
        \cline{5-6}
        & & & & \hspace{2.5em} VGG-11: $36.800$ & $4.600{\times}10^{3}\times$ \\
        \hline
        \multirow{2}{*}{CIFAR-100} & \multirow{2}{*}{20} & \multirow{2}{*}{100}
            & \multirow{2}{*}{$0.0080$}
            & \hspace{2.1em} LeNet-5: $0.250$ & \hspace{2em} $31.25\times$ \\
        \cline{5-6}
        & & & & \hspace{2.5em} VGG-11: $36.821$ & $4.603{\times}10^{3}\times$ \\
        \hline
        \multirow{2}{*}{PathMNIST} & \multirow{2}{*}{9} & \multirow{2}{*}{100}
            & \multirow{2}{*}{$0.0036$}
            & EfficientNet-B0: $21.195$ & $5.888{\times}10^{3}\times$ \\
        \cline{5-6}
        & & & & EfficientNet-B3: $47.994$ & $1.333{\times}10^{4}\times$ \\
        \hline
        \multirow{2}{*}{TissueMNIST} & \multirow{2}{*}{8} & \multirow{2}{*}{100}
            & \multirow{2}{*}{$0.0032$}
            & EfficientNet-B0: $21.190$ & $6.622{\times}10^{3}\times$ \\
        \cline{5-6}
        & & & & EfficientNet-B3: $47.988$ & $1.500{\times}10^{4}\times$ \\
        \hline
        \multirow{2}{*}{Brain-Tumor MRI} & \multirow{2}{*}{4} & \multirow{2}{*}{40}
            & \multirow{2}{*}{$0.00064$}
            & EfficientNet-B0: $21.169$ & $3.308{\times}10^{4}\times$ \\
        \cline{5-6}
        & & & & EfficientNet-B3: $47.963$ & $7.494{\times}10^{4}\times$ \\
        \hline
        \end{tabular}
    }
    \label{tab:comm-totals}
\end{table}

Thus, FedEntOpt has nearly the same per-round communication as FedAvg, since it adds only a negligible one-time cost before training. The communication cost conclusions are identical when applying differential privacy via the Laplace mechanism according to \eqref{eq:laplace}, since each client still transmits $C$ 32-bit scalars.

\subsection{Buffer Size and Selection Diversity}
\label{sec:ablation-q-seed}
To study how the buffer size affects the diversity in client selection, we rely on the CIFAR-10 and CIFAR-100 datasets under the standard settings described in the hyperparameter section,  using $C{=}2$ and $\mathrm{Dir}(0.1)$ as partitioning strategies.  
To quantify diversity, we count how many times each client is selected across all communication rounds, normalize the counts into a probability distribution, and compute its normalized entropy $\mathbb{H}_{\text{norm}}$. It is defined as
\[
    \mathbb{H}_{\text{norm}}:= \frac{-\sum_{k=1}^{K} p_k \log_2 p_k}{\log_2 K} \in [0,1],
    \quad p_k=\frac{c_k}{\sum_{j=1}^{K} c_j},
\]
where $c_k$ is the number of times client $k$ was selected. We vary the buffer sizes $Q \in \{0\%,\,25\%,\,50\%,\,75\%\}$ of the
available clients. To assess the effect on model performance across these $Q$ values, we utilize the LeNet-5 model. Results are reported in Table~\ref{tab:ablation-q-entropy-cifar} as mean with standard deviation over 3 random seeds.

\begin{table}[ht]
    \small
    \centering
    \caption{Effect of buffer size $Q$ (\% of available clients) on CIFAR-10 and CIFAR-100 with LeNet-5 architecture. We report mean test accuracy (\%) over the final 10 rounds of FedEntOpt and normalized selection entropy $\mathbb{H}_{\text{norm}}$ (higher is better, indicating more diverse client participation) under $\mathrm{Dir}(0.1)$ and $C=2$.}
    \setlength{\tabcolsep}{1mm}
    \resizebox{\columnwidth}{!}{%
        \renewcommand{\arraystretch}{1.35}
        \begin{tabular}{|c|c|c|c|c|}
        \hline
        \textbf{Dataset} & \textbf{Partition} & \textbf{Buffer Size Q} & \textbf{FedEntOpt} & \textbf{$\mathbb{H}_{\text{norm}}$} \\
        \hline
        \multirow{8}{*}{CIFAR-10}
            & \multirow{4}{*}{$C=2$}
                & $0$   & $49.34 \pm 0.82$ & $0.715 \pm 0.006$ \\
            \cline{3-5}
            &   & $25$  & $53.25 \pm 0.20$ & $0.918 \pm 0.003$ \\
            \cline{3-5}
            &   & $50$  & $53.39 \pm 0.18$ & $0.976 \pm 0.001$ \\
            \cline{3-5}
            &   & $75$  & $53.82 \pm 0.09$ & $0.998 \pm 0.001$ \\
        \cline{2-5}
            & \multirow{4}{*}{$\mathrm{Dir}(0.1)$}
                & $0$   & $49.84 \pm 0.77$ & $0.896 \pm 0.002$ \\
            \cline{3-5}
            &   & $25$  & $52.55 \pm 0.23$ & $0.947 \pm 0.006$ \\
            \cline{3-5}
            &   & $50$  & $52.92 \pm 0.11$ & $0.979 \pm 0.002$ \\
            \cline{3-5}
            &   & $75$  & $52.64 \pm 0.08$ & $0.998 \pm 0.001$ \\
        \hline
        \multirow{8}{*}{CIFAR-100}
            & \multirow{4}{*}{$C=2$}
                & $0$   & \multicolumn{1}{c|}{$26.21 \pm 0.86$} & \multicolumn{1}{c|}{$0.818 \pm 0.010$} \\
            \cline{3-5}
            &   & $25$  & \multicolumn{1}{c|}{$27.33 \pm 0.35$} & \multicolumn{1}{c|}{$0.945 \pm 0.005$} \\
            \cline{3-5}
            &   & $50$  & \multicolumn{1}{c|}{$27.60 \pm 0.11$} & \multicolumn{1}{c|}{$0.982 \pm 0.002$} \\
            \cline{3-5}
            &   & $75$  & \multicolumn{1}{c|}{$27.96 \pm 0.19$} & \multicolumn{1}{c|}{$0.998 \pm 0.002$} \\
        \cline{2-5}
            & \multirow{4}{*}{$\mathrm{Dir}(0.1)$}
                & $0$   & $32.62 \pm 0.71 $ & $0.853 \pm 0.005$ \\
            \cline{3-5}
            &   & $25$  & $33.07 \pm 0.41 $ & $0.933 \pm 0.004$ \\
            \cline{3-5}
            &   & $50$  & $32.72 \pm 0.13 $ & $0.970 \pm 0.002$ \\
            \cline{3-5}
            &   & $75$  & $32.98 \pm 0.18$ & $0.993 \pm 0.001$ \\
        \hline
        \end{tabular}
    }
    \label{tab:ablation-q-entropy-cifar}
\end{table}

Across both partitioning schemes and datasets, increasing the buffer size consistently raised the normalized entropy score, indicating more uniform and diverse client participation. However, the impact on model performance differs by the partitioning strategy. Under $\mathrm{Dir}(0.1)$, labels are concentrated among a few clients. Larger buffer size excludes these clients, degrading the global model performance. In contrast, for $C=2$, where each label is widely represented across clients, larger buffer sizes consistently improve performance by preventing FedEntOpt to become biased towards a smaller cohort of clients.
Note that in all settings, every client was selected at least once.

\section{Discussion and Limitations}
FedEntOpt is a round-wise greedy client selection method that maximizes the entropy of the aggregated label distribution, keeping the procedure lightweight and efficient. While effective in our settings, it could potentially underperform in cases where label availability is extremely sparse, i.e., clients may hold near single-class datasets, and the participation is very low. In such cases, the entropy of the aggregated subset can plateau, leaving some classes underrepresented. A practical mitigation is to increase the FIFO buffer size to improve the selection diversity. FedEntOpt also relies on the fidelity of the uploaded label counts. While our results indicate robustness to moderate differential privacy noise, adversarial clients could still misguide the client selection by transmitting highly noisy label counts or mislabeled data.
In practice, limited participation and a larger buffer size could reduce any single adversarial client’s influence, but dedicated defenses are needed against strong adversaries.

\section{Conclusion}
In this paper, we proposed a new client selection method, named FedEntOpt, which utilizes label information of individual clients to mitigate the influence of label skew in federated learning systems. By iteratively maximizing the entropy of the aggregated label distribution of selected clients, the global model benefits from receiving updates from a complementary set of clients. Extensive experiments showed that the approach significantly outperformed other state-of-the-art methods in terms of classification accuracy regardless of the model size. Moreover, FedEntOpt consistently delivers superior results not only on low-resolution benchmark datasets but also, when applied to higher-resolution, real-world data.
Furthermore, FedEntOpt demonstrates robust performance in challenging scenarios characterized by low client participation rates, client dropouts, and straggling, while ensuring strong privacy guarantees. Due to its simplicity, it is readily integrated with other methods to significantly enhance their performance.

\begin{acronym}
    \acro{fl}[FL]{federated learning}
    \acro{fedavg}[FedAvg]{federated averaging}
\end{acronym}

\bibliographystyle{IEEEtran}
\bibliography{references}

\end{document}